\pdfoutput=1

\documentclass[11pt]{article}

\usepackage[final]{style/acl}

\usepackage{times}
\usepackage{latexsym}

\usepackage[T1]{fontenc}

\usepackage[utf8]{inputenc}

\usepackage{microtype}

\usepackage{inconsolata}

\usepackage{graphicx}

\usepackage{amssymb}
\usepackage{amsmath}
\usepackage{subcaption}
\usepackage{booktabs}
\usepackage{xcolor}
\usepackage{multirow}
\usepackage{colortbl}
\title{\speechQE{}: Estimating the Quality of Direct Speech Translation}

\author{HyoJung Han \\
Computer Science\\
  University of Maryland \\
  \texttt{hjhan@cs.umd.edu} \\
  \And
  Kevin Duh \\
  HLTCOE\\
  Johns Hopkins University \\
  \texttt{kevinduh@cs.jhu.edu} \\
  \And
  Marine Carpuat \\
Computer Science\\
  University of Maryland \\
  \texttt{marine@cs.umd.edu} \\
  }

\newcommand{\abr}[1]{\textsc{#1}}
\newcommand{\textQE}[0]{text-QE}
\newcommand{\textLLM}[0]{text-LLM}
\newcommand{\speechQE}[0]{SpeechQE}
\newcommand{\textESD}[0]{text-ESD}
\newcommand{\speechESD}[0]{SpeechESD}

\begin{document}
\maketitle
\begin{abstract}
Recent advances in automatic quality estimation for machine translation have exclusively focused on written language, leaving the speech modality underexplored.
In this work, we formulate the task of quality estimation for speech translation, construct a benchmark, and evaluate a family of systems based on cascaded and end-to-end architectures. In this process, we introduce a novel end-to-end system leveraging pre-trained text LLM. 
Results suggest that end-to-end approaches are better suited to estimating the quality of direct speech translation than using quality estimation systems designed for text in cascaded systems.
More broadly, we argue that quality estimation of speech translation needs to be studied as a separate problem from that of text, and release our data and models to guide further research in this space.\footnote{\url{https://github.com/h-j-han/SpeechQE}}

\end{abstract}
\section{Introduction}
\label{sec:intro}

Recent progress in quality estimation (QE) \cite{Specia2010} makes it possible to automatically rate the quality of machine translation (MT) given only the input and output of an MT system. QE ratings have been found to correlate well with human judgments, sometimes as well as reference-based metrics \citep{kepler-etal-2019-openkiwi, rei-etal-2020-comet, rei-etal-2023-scaling}. However, this work has focused on text translation.

Meanwhile, the rapid development of speech technology \citep{radford2022robust,communication2023seamlessm4t} has expanded the use of speech translation (ST) applications in daily life, thus increasing the need to predict the reliability of their output.
This raises the question of whether quality estimation for ST can be performed using a combination of state-of-the-art automatic speech recognition (ASR) and text-based QE (\textQE{} or MTQE) methods.
However, relying on a cascade of ASR and \textQE{} systems presents two major issues: 
(1) The current top-performing ST models directly translate the audio input into target language text without transcribing the audio, making it inefficient to run an additional ASR system to generate an input for the \textQE{} module.
(2) ASR transcriptions of the audio input may not match the gold transcription, potentially misleading the \textQE{} system.
Hence, we hypothesize that end-to-end approaches might be better suited for this task.

\begin{figure}[t]
    \centering
    \includegraphics[width=0.48\textwidth]{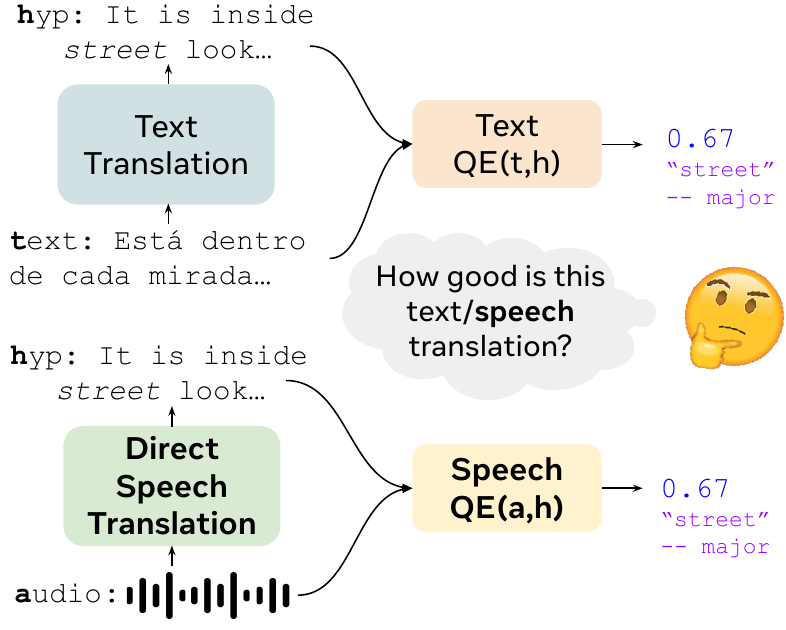} 
    \caption{Quality Estimation for Speech Translation (\speechQE{}) vs. Text Quality Estimation (\textQE{}).}
    \label{fig:task_text_speech}
\end{figure}

\begin{figure*}[t]
    \centering
    \includegraphics[width=0.88\textwidth]{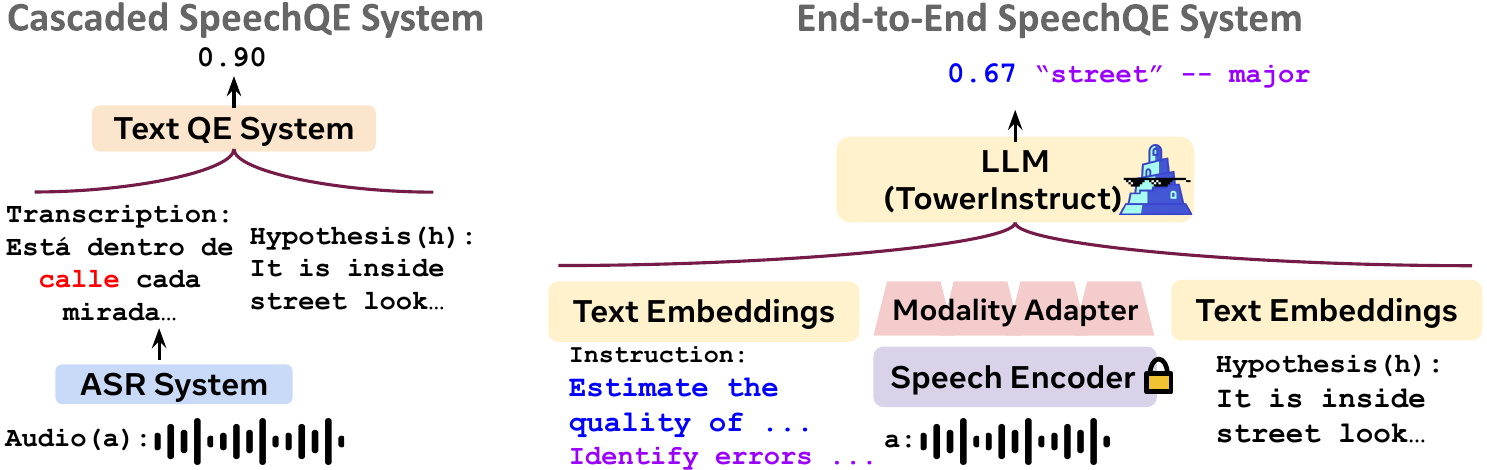} 
    \caption{Comparing cascaded and end-to-end approaches to Quality Estimation for Speech Translation (\speechQE{}).
    }
    \label{fig:task_e2e_cas}
\end{figure*}

In light of these issues, we formulate the task of \textbf{quality estimation for speech translation (\speechQE{}} or STQE, Figure~\ref{fig:task_text_speech}) and explore both cascaded and end-to-end (E2E) systems for this task (Figure~\ref{fig:task_e2e_cas}).
While we rely on existing ASR and \textQE{} modules for the cascaded system, 
we introduce a novel E2E \speechQE{} model architecture to address the lack of a dedicated end-to-end system for this task.  Our design incorporates a pre-trained speech encoder and a large language model (LLM) to leverage their existing capabilities in extracting high-quality audio features and handling translation-related tasks. 

To conduct a thorough evaluation, we contribute an evaluation benchmark and training data for \speechQE{} from diverse ST outputs scored with reference-based metrics.
%
Results show that E2E models outperform the cascaded system based on a state-of-the-art (SOTA) ASR module in correlation with both (1) human direct assessment ratings and (2) metric scores. 
Additionally, our E2E model can detect error spans to some extent in a zero-shot fashion, though the best results are still achieved by cascaded systems with SOTA ASR.
Qualitative analysis highlights the robustness of E2E models against wrong speech representation in score prediction, error span detection, and severity prediction. 
Based on this evidence, we argue that \speechQE{} should be studied as a distinct problem from \textQE{}.

\section{Background} 
\label{sec:background}
Quality estimation makes it possible to assess translation quality without reference translations, which is essential for practical use cases \citep{Specia2010,callison-burch-etal-2012-findings}. 
QE signals can benefit end users by helping them decide how to rely on outputs in casual and high-risk settings alike \citep{specia2022quality, mehandru-etal-2023-physician}. They can also benefit downstream tasks or enhance MT itself \citep{fernandes-etal-2022-quality}.

The QE task has been framed in various ways, including predicting sentence-level quality ratings \citep{callison-burch-etal-2012-findings} or word-level binary tags of OK/BAD \citep{bojar-etal-2013-findings}.
While a wealth of methods have been developed for these tasks,  recent work has shown the benefits of developing solutions to address them jointly. OpenKiwi \citep{kepler-etal-2019-openkiwi} streamlined QE by supporting both word-level tagging and regression toward a sentence-level score within a unified toolkit \citep{kim-etal-2017-predictor}. It was further improved with a training recipe that better supports multilingual generalization \citep{rei-etal-2020-comet, rei-etal-2023-scaling}.
Together with the development of learned metrics for reference-based evaluation \cite{rei-etal-2020-comet,sellam-etal-2020-bleurt}, this set the stage for a single or family of models that flexibly rate the quality of MT output with or without access to a reference human translation \citep{guerreiro2023xcomet, juraska-etal-2023-metricx} with high correlations with human quality ratings \citep{freitag-etal-2023-results}. xCOMET~\citep{guerreiro2023xcomet} even integrates both sentence-level evaluation and error span detection capabilities while categorizing error spans, thereby enriching the quality measures.

Meanwhile, quality estimation for speech translation remains understudied.  \citet{le-etal-2016-joint} 
address the task of tagging each word in an ST output as good or bad, using ASR and MT features. Their approach can be viewed as a cascaded SpeechQE system, which propagates a confidence score in a pipeline of ASR and statistical machine translation (SMT) modules. 
BLASER2.0~\citep{communication2023seamlessm4t} produces a similarity score between a translation output and input, using SONAR sentence-embeddings that can compare either speech or text \cite{sonar23}. While this enables \speechQE{}, this approach was initially designed for speech-to-speech translation~\citep{chen-etal-2023-blaser}, and was exposed to only a small amount of training data with quality labels.

With advances in ST technology and their growing use \cite{rubenstein2023audiopalm}, 
there is a need for QE to support ST scenarios where intermediate automatic speech recognition (ASR) outputs are not available, along with new evaluations to correctly gauge the effectiveness of quality estimation in speech translation.


\section{\speechQE{}: Task and Models}
\label{sec:methods}

We define the task of estimating the quality of speech translation (\speechQE{} or STQE\footnote{We choose to use terms \speechQE{} and \textQE{} as main instead of alternative terms STQE and MTQE to emphasize the contrast between speech and text and to facilitate easier reading. More discussion of the terminology in Appendix~\ref{sec:task_name}.}), before introducing our cascaded and E2E systems. 

In this work, we focus on predicting sentence-level scores and measuring the correlation of reference ratings provided by humans or reference-based metrics \citep{fonseca-etal-2019-findings}.
Additionally, we will explore an error span detection task \citep{blain-etal-2023-findings} in Section~\ref{sec:results_esd4st}, to broaden the scope of QE beyond holistic numerical ratings.

We refer to a \textbf{reference-based metric} as $metric$. Given a reference target text $r$, an MT hypothesis $h$ and optionally the MT source text $t$, the $metric$ rates the quality of $h$ as a score $m$:
\begin{equation} m = metric(h, r)\text{  or  }m = metric(t, h, r) \end{equation}
Likewise, we refer to a \textbf{text quality estimation} system as $\text{\textQE{}}$. It produces an output score $q$ given only a source text $t$ and an MT hypothesis $h$.
\begin{equation} q = \text{\textQE{}}(t, h)\end{equation}
In the \textbf{\speechQE{} task} (Figure~\ref{fig:task_text_speech}), given the source \textbf{audio} $a$ and the translation hypothesis $h$, a system outputs the quality score $q$ for this hypothesis:
\begin{equation} q = \speechQE{}(a, h)\end{equation}

\subsection{Cascaded \speechQE{} System}
\label{sec:methods_cas}

We first consider cascaded \speechQE{} systems that output the score $q_{cas}$ from a text-based QE system with the input of transcribed text $ASR(a)$ from an ASR system and hypothesis text $h$ (Figure~\ref{fig:task_e2e_cas}).
\begin{equation} q_{cas} = \text{\textQE{}}(ASR(a), h)\end{equation}

While the cascaded systems offer a straightforward approach to \speechQE{}, they present several issues.  
First, efficiency is a concern, as 
there are no naturally occurring intermediate ASR transcripts in the case of direct ST, necessitating additional ASR runs to generate inputs for the \textQE{} component.
This introduces latency that may be undesirable in user-facing quality estimation applications. 
Second, 
source transcriptions produced by a separate ASR do not always accurately represent the spoken input, making the \textQE{} system vulnerable to the wrong speech representation.
Third, there is a modality mismatch, as the \textQE{} component is not adapted to spoken language, which exhibits different styles or errors from written language. 
These challenges motivate us to explore end-to-end (E2E) \speechQE{} solutions.

\subsection{End-to-End \speechQE{} System}
\label{sec:methods_e2e}

We introduce the architecture and training scheme for our E2E \speechQE{} model. 

\paragraph{Model Architecture}

Rather than training an integrated model from scratch, we choose to leverage a pre-trained speech encoder and a large language model (LLM) to utilize their abilities in extracting high-quality audio features and handling translation-related tasks, respectively. This approach is particularly useful when there is limited or no data available for training from scratch, as it enables the transfer of knowledge from text-based large language models (\textLLM{}) to the speech domain. 
We adopt a popular configuration for integrating speech modality into \textLLM{} that trains a lightweight modality adapter \citep{wu2023decoderonly, fathullah2023prompting, wang2023blsp, wang2023slm}, but the optimal architecture for \speechQE{} or even broadly for integrating speech modality into text language model remains an open question.

Figure~\ref{fig:task_e2e_cas} shows the overview of E2E system architecture.
The E2E \speechQE{} model has three parts: pre-trained speech encoder, modality adapter, and pre-trained \textLLM{}. 
The speech encoder extracts the audio feature from the raw audio, where we initialize with existing competitive speech models.
The modality adapter subsamples the audio features to compress the audio sequence and bridges the speech representation to the text embedding space to output speech embeddings.
We fix the speech encoder for all experiments, while the weights of the adapter and \textLLM{} can be updated depending on the training settings.
The input of the \textLLM{} model is the concatenation of text and audio embedding sequence.

\paragraph{Training}

Supervised \speechQE{} training and evaluation requires triplets of audio inputs, ST hypotheses, and quality ratings. We build a corpus by generating hypotheses with direct ST systems of varying quality and obtain automatic quality labels from reference-based metric (\textsection~\ref{sec:exp_building_corpus}).\footnote{
This is intended to minimize any bias from the written text domain, rather than augment speech modality with TTS on existing text datasets with human scores.}
We train the E2E model with the \speechQE{} task, complemented with the ASR and ST tasks which provide supervision of mapping between text and speech modality. 
We consider two training strategies.
The first is a simple single-phase approach where we train a modality adapter (and optionally update \textLLM{}) with all three tasks.
The second is a two-phase approach where we first train only an adapter with ASR and ST tasks while freezing \textLLM{} to focus solely on mapping between text and speech modality. Then, we continue training with the \speechQE{} task to let the LLM learn the unseen task of QE. 
In the second phase, the adapter pre-trained in the previous phase can be frozen or updated, while \textLLM{} is always trained with LoRA \citep{hu2022lora}.

We now turn to the empirical evaluation to determine whether the E2E model successfully overcomes the efficiency and modality alignment issues raised by cascaded systems. 



\section{Experimental Settings}
\label{sec:exp}

\begin{table}[t]
\centering
\begin{tabular}{crrr}
\toprule
CoVoST2/CV4       & ASR  & ST   & \speechQE{} \\
\midrule
es2en & 297k & 79k  & 546k \\
en2de & 305k & 290k & 589k \\\bottomrule
\end{tabular}
\caption{Number of instances of training corpus of each speech related tasks. CoVoST2 for ST and \speechQE{}, and Common Voice 4 for ASR. \speechQE{} set is generated from the subset of ST by seven translation systems.}
\label{table:data_statistics}
\end{table}
\begin{table}[]
\centering
\fontsize{8.5}{13}\selectfont
\begin{tabular}{lll}
\toprule
Es2En diect ST systems      & CoVoST2 & FLEURS \\
\midrule
whisper-large-v3 & 39.05   & 22.45  \\
whisper-large-v2 & 39.53   & 23.62  \\
whisper-large    & 38.11   & 22.89  \\
whisper-medium   & 37.39   & 21.93  \\
whisper-small    & 31.27   & 17.78  \\
whisper-base     & 16.93   & 11.67  \\
whisper-tiny     & 7.81    & 6.86  \\ 
\toprule
En2De direct ST systems                      & CoVoST2 & FLEURS \\
\midrule
seamless-m4t-v2-large            & 43.12   & 32.21  \\
seamless-m4t-large               & 40.55   & 31.41  \\
seamless-m4t-medium              & 38.39   & 26.83  \\
s2t-wav2vec2-large-en-de         & 26.98   & 19.92  \\
s2t-medium-mustc-multilingual-st & 8.08    & 13.43  \\
s2t-small-mustc-en-de-st         & 7.82    & 12.34  \\
s2t-small-covost2-en-de-st       & 14.19   & 9.50    \\\bottomrule
\end{tabular}
\caption{The list of seven direct ST models and their BLEU scores for generating training corpus and test benchmarks of \speechQE{}.}
\label{table:bleu}
\end{table} 

In this section, we describe the construction of the \speechQE{} benchmark as well as the configuration of the evaluated systems.

\subsection{Building \speechQE{} Benchmark}
\label{sec:exp_building_corpus}

\begin{table*}[t]
\centering
\fontsize{9.0}{13}\selectfont
\begin{tabular}{lcccc}
\toprule
$\rho = corr(\text{\textbf{q}}, \text{\textbf{m}})$ \quad\quad\quad\quad$\text{m}_{\text{xCOMET}} = \text{xCOMET}(\text{gold }t,h,r)$ & \multicolumn{2}{c}{Es2En} & \multicolumn{2}{c}{En2De} \\
\quad\quad\quad\quad\quad\quad\quad\quad\quad\quad\quad\quad\quad\quad$\,\,\text{m}_{\text{MetricX}} = \text{MetricX}(h,r)$  & $\text{m}_{\text{xCOMET}}$      & $\text{m}_{\text{MetricX}}$     & $\text{m}_{\text{xCOMET}}$      & $\text{m}_{\text{MetricX}}$     \\
\midrule[\heavyrulewidth]
\multicolumn{5}{c}{\textit{\textbf{Cascaded \speechQE{} Systems Correlations}}$\,\rho_{cas} = corr(\text{\textbf{q}}_{cas}, \text{\textbf{m}})$}                                                                                                               \\
$\text{q}_{cas} = \text{xCOMET-qe}(\text{gold }t,h)$  & 0.929  & $\overline{\mbox{0.812}}$  & 0.967  & $\overline{\mbox{0.872}}$  \\
$\text{q}_{cas} = \text{xCOMET-qe}(\text{ASR}(a),h)$ & $\overline{\mbox{0.892}}$  & 0.782  & $\overline{\mbox{0.910}}$  & 0.821  \\\hline
$\text{q}_{cas} = \text{MetricX-qe}(\text{gold }t,h)$ & $\overline{\mbox{0.834}}$ & \underline{0.844}   & $\overline{\mbox{0.908}}$ & 0.932   \\
$\text{q}_{cas} = \text{MetricX-qe}(\text{ASR}(a),h)$ & 0.803 & 0.803   & 0.854 & $\overline{\mbox{0.871}}$   \\
\hline
$\text{q}_{cas} = \text{text-BLASER2.0-qe}(\text{gold }t,h)$ & $\overline{\mbox{0.813}}$ &	$\overline{\mbox{0.739}}$&	$\overline{\mbox{0.870}}$&	$\overline{\mbox{0.833}}$   \\
$\text{q}_{cas} = \text{text-BLASER2.0-qe}(\text{ASR}(a),h)$ & 0.776 & 0.711   & 0.813 & 0.771   \\
\midrule[\heavyrulewidth]
\multicolumn{5}{c}{\textit{\textbf{End-to-End \speechQE{} Systems Correlations}}$\,\rho_{e2e} = corr(\text{\textbf{q}}_{e2e}, \text{\textbf{m}})$}                                                                                                             \\
$\text{q}_{e2e} = \text{BLASER2.0-qe}(a,h)$  & 0.780    & 0.712     & 0.856    & 0.819     \\
\hline
$\text{q}_{e2e} = \text{\textit{TowerInstruct-Fixed+Adapter}}(a,h)$  & 0.862    & 0.797     & 0.882    & 0.848     \\
$\text{q}_{e2e} = \text{\textit{TowerInstruct-LoRA+Adapter}}(a,h)$   & 0.882    & 0.818     & 0.914    & 0.867     \\
$\text{q}_{e2e} = \text{\textit{TowerInstruct-LoRA+Adapter-pt}}(a,h)$      & 0.890  & 0.833  & 0.922  & 0.872  \\
$\text{q}_{e2e} = \text{\textit{TowerInstruct-LoRA+Adapter-pt-Fixed}}(a,h)$ & \textbf{0.895}  & \textbf{0.834}  & \textbf{0.925}  &\textbf{ 0.873}               \\\bottomrule
\end{tabular}
\caption{Correlations ($\rho$) between \speechQE{} system scores (\textbf{q}) and metric scores (\textbf{m}) for quality of ST on CoVoST2 test.
ASR is whisper-large-v3, the cutting-edge model.
E2E systems outperform ASR cascaded systems and even some cascaded ones with gold transcriptions.
Overlines in cascaded correlation mean that the best E2E system outperforms the corresponding cascaded system. 
Bolded text in E2E indicate the best score within each column.}
\label{table:cas_vs_e2e_indomain}
\end{table*}

We build a training corpus and test benchmark for \speechQE{} from CoVoST2 \citep{wang21s_interspeech} which is a speech translation corpus based on Common Voice 4 ASR datasets \citep{ardila-etal-2020-common}.
We consider two translation directions: Spanish-to-English and English-to-German.
We subsample about 80k segments from the training set and 500 from the dev and test of CoVoST2, then run seven different direct ST models to generate the ST hypotheses.
The direct ST models are off-the-shelf models of a wide range of translation quality including Whisper \citep{radford2022whisper} for Es2En, and Seamless-M4T \citep{communication2023seamlessm4t} and Fairseq S2T \citep{wang-etal-2020-fairseq} for En2De.
The details of ST models are in Table~\ref{table:bleu}.

Given the generated hypothesis text, reference text, and gold transcription text, we get automatic quality labels from (reference-based) metrics since reference-based scores are generally known to be better correlated with human judgment on translation quality than reference-free scores \citep{freitag-etal-2023-results}.
For training, we choose xCOMET-XL \citep{guerreiro2023xcomet} as metric because it is one of the best-performing submissions in the WMT23 metric shared task.
The final statistics for the training dataset are in Table~\ref{table:data_statistics}.
For the test, we obtain metric scores from both xCOMET-XL and MetricX-23-XL \citep{juraska-etal-2023-metricx} 
as two distinct types of quality labels to avoid biased comparison with the cascaded system.

\subsection{Cascaded Modeling}
\label{sec:exp_cas}
For the cascaded system, we use the same set of Whisper models that generates the Es2En ST hypothesis as the ASR module for both the Es2En and En2De cascaded experiments. 
For QE modules, we use the same metric models that generate reference-based quality labels in Section~\ref{sec:exp_building_corpus} but with reference-free inputs: source and hyothesis.\footnote{We choose to report QE decoding of  MetricX-23-XL instead of the dedicated QE model of MetricX-23-QE-XL as the former has higher correlations with human DA and the findings in the Results sections are the same.}

\subsection{E2E Modeling}
\label{sec:exp_e2e}

We initialize the speech encoder from Whisper-large-v2 and freeze it for all experiments. The \textLLM{} is TowerInstruct-7B \citep{tower_llm_2024} which is continued pre-training and finetuned with instructions relevant to translation processes from Llama 2 \citep{touvron2023llama}.
This model has not trained on the task of predicting the quality score of a given translation (QE) but has trained on the error span detection task.
We either freeze the TowerInstuct model or train it with LoRA ($r=16$, $\alpha=32$).
The modality adapter consists of three 1-dimensional convolutional layers followed by a 512-dimensional bottleneck layer \citep{pmlr-v97-houlsby19a}, following \citet{wang2023blsp}.
The adapter is initialized randomly and unfrozen unless stated. 

All our E2E models are trained on a single A6000 GPU with a batch size of 8 updated in fixed steps (140k steps for the single phase strategy, and 120k+80k steps for the two-phase strategy).
In addition to the \speechQE{} training set, we use Common Voice 4 and CoVoST2 for ASR and ST. 
We use language modeling loss with fixed instruction prompts for each task for all settings, following the chat template of TowerInstruct. 
More experimental details are in Appendix~\ref{sec:add_exp_detail} including the instruction prompt templates for each task (Figure~\ref{fig:prompt_exmaple}).

As another baseline, we use the BLASER2.0-qe to experiment with both cascaded and E2E scenarios. 
The inputs of E2E setting are SONAR embedding of source speech and target text, while all text embedding is for the cascaded setting. 

\subsection{Evaluation}
\label{sec:exp_eval}

We evaluate all models on the \speechQE{} test set built in Section~\ref{sec:exp_building_corpus}, which has two types of metric labels from xCOMET-XL and MetricX-XL.
A lower score of MetricX indicates better quality, while that of xCOMET and E2E systems indicates the opposite. 
To simplify our analysis, we multiply MetricX scores by negative one, which allows us to focus on the extent of correlation without considering the direction.
We use the Spearman as the primary measurement following \citet{blain-etal-2023-findings}.

For evaluation on quality labels by human judgement instead of metric, we compare human direct assessment (DA) score on IWSLT ACL set from \citet{sperber-etal-2024-evaluating-iwslt2023} which is based on \citet{salesky-etal-2023-evaluating}.\footnote{\url{https://huggingface.co/datasets/IWSLT/da2023}}
This dataset is based on presentation videos describing their ACL papers, thus including highly technical terms and having domain mismatches between our main training corpus. 
It contains the source-based DA ratings of 416 hypotheses from each of the ten ST systems, resulting in a total of 4,160 instances.
We include additional QE and metric models including sentence BLEU and Comet(KiWi) \citep{ rei-etal-2022-comet, rei-etal-2022-cometkiwi, rei-etal-2023-scaling}.


\section{Results}
\label{sec:results}

We first present our main results by comparing \speechQE{} ratings with reference-based metrics (\textsection~\ref{sec:results_main}), then turn to using human ratings of translation quality (\textsection~\ref{sec:results_humanda}). 
We add the results of varying model sizes and architecture of the cascaded system. (\textsection~\ref{sec:results_size}).
Finally, we evaluate our models on a zero-shot error detection task (\textsection~\ref{sec:results_esd4st}) and conduct a qualitative analysis of outputs (\textsection~\ref{sec:results_example}).
We additionally evaluate and train our systems with out-of-domain settings (Appendix \ref{sec:results_ood} and \ref{sec:results_add_fleurs_train}). 

\subsection{Correlation with Reference-based Metrics}
\label{sec:results_main}

Table~\ref{table:cas_vs_e2e_indomain} shows correlations between metric scores as quality labels and \speechQE{} system output scores, where the input of metric includes gold transcription source text and reference text. 

\paragraph{Cascaded.} 
For metric and \textQE{} scores, we cross-compare two metric scores (xCOMET and MetricX) as quality labels and two QE scores (xCOMET-qe and MetricX-qe) within cascaded configurations since the matching QE and metric model could favor the output from the model similar to its own.
For example, the xCOMET is a single model for both metric and QE with different inputs, showing higher correlation values in the metric-QE model matching configuration (0.929 in Es2En) than mismatch (0.834 or 0.812).

\paragraph{E2E.} 
Among four E2E models, LoRA training the \textLLM{} with a fixed pre-trained speech adapter (\textit{TowerInstruct-LoRA+Adapter-pt-Fixed}) performs the best in all language pairs and metric types. 
The simplest training of fixing LLM and updating only the adapter with all three tasks in a single phase (\textit{TowerInstruct-Fixed+Adapter}) shows the lowest correlations followed by similar methods but LoRA training the \textLLM{} (\textit{TowerInstruct-LoRA+Adapter}).
This suggests that a separate training phase for mapping speech-to-text perception is critical and that the weight updates are necessary when a \textLLM{} is not fine-tuned for the target task and therefore lacks the required capabilities.
In this case, TowerInstruct is not fine-tuned with QE tasks, therefore, updating it is necessary. 
All variants of our E2E system outperform BLASER2.0, perhaps due to its limited exposure to diverse translation quality at training time.

\paragraph{E2E vs Cascaded.} 
The end-to-end \speechQE{} systems consistently outperform the cascaded system which included the SOTA ASR system (whisper-large-v3).  The best E2E system not only outperforms ASR-based cascades, but cascaded systems that use gold transcriptions in all QE(row)-metric(column) mismatched settings of both language pairs.
For instance, 0.834 of E2E versus 0.812 of xCOMET-qe(gold $t$,$h$) cascaded in Es2En MetricX column.
Similarly, BLASER2.0 with the E2E setting of speech input and text output outperforms the cascade system with the text input-output setting (text-BLASER2.0).

Overall, the correlation analysis underscores the advantage of end-to-end \speechQE{} systems over cascaded ones. 
The strong correlations with metric scores across various configurations indicate its reliability as a measurement for quality estimation in automatic speech translation tasks, highlighting the potential of end-to-end approaches.

\subsection{\speechQE{} Correlation with Human DA}
 \label{sec:results_humanda}

\begin{table}[t]
\centering
\fontsize{8.2}{12}\selectfont
\begin{tabular}{lc}
\toprule
IWSLT23-ACL En2De Test set &  Human DA \\
$\rho = corr(\text{\textbf{x}}, \text{\textbf{d}})$   &  score $d$ \\
\midrule[\heavyrulewidth]
\multicolumn{2}{l}{\textit{\textbf{Metric} and Human DA correlation }$\rho = corr(\text{\textbf{m}}, \text{\textbf{d}})$}          \\
$\text{m}\text{ = }\text{xCOMET}(\text{gold }t,h,r)$       & \textbf{0.557}        \\
$\text{m}\text{ = }\text{MetricX}(h,r)$      & 0.539       \\
$\text{m}\text{ = }\text{wmt22-comet-da}(\text{gold }t,h,r)$                                             & 0.544        \\
$\text{m}\text{ = }\text{sentBLEU}(h,r)$                                              & 0.336        \\
\midrule[\heavyrulewidth]
\multicolumn{2}{l}{\textit{\textbf{Cascaded \speechQE{}} and Human DA }$\rho = corr(\text{\textbf{q}}_{cas}, \text{\textbf{d}})$}  \\
$q\text{ = }\text{xCOMET-qe}(\text{gold }t,h)$                    & 0.544       \\
$q\text{ = }\text{MetricX-qe}(\text{gold }t,h)$                   & 0.556      \\
$q\text{ = }\text{wmt23-cometkiwi-da-xl}(\text{gold }t,h)$        & 0.576       \\
$q\text{ = }\text{wmt22-cometkiwi-da}(\text{gold }t,h)$          & \textbf{0.580}       \\
\hline
$q\text{ = }\text{xCOMET-qe}(\text{ASR}(a),h)$             & 0.485       \\
$q\text{ = }\text{MetricX-qe}(\text{ASR}(a),h)$            & 0.495      \\
$q\text{ = }\text{wmt23-cometkiwi-da-xl}(\text{ASR}(a),h)$ & \textbf{0.503}       \\
$q\text{ = }\text{wmt22-cometkiwi-da}(\text{ASR}(a),h)$    & 0.486      \\
$q\text{ = }\text{text-BLASER2.0-qe}(\text{ASR}(a),h)$    & 0.428      \\
\midrule[\heavyrulewidth]
\multicolumn{2}{l}{\textit{\textbf{E2E \speechQE{}} \& Human DA correlation }$\rho = corr(\text{\textbf{q}}_{e2e}, \text{\textbf{d}})$}       \\
$q\text{ = }\text{BLASER2.0-qe}(a,h)$       & 0.420       \\
$q\text{ = }\text{\textit{TowerInst-LoRA+Adapter-pt}}(a,h)$       & 0.492       \\
$q\text{ = }\text{\textit{TowerInst-LoRA+Adapter-pt-Fixed}}(a,h)$ & \textbf{0.509}    \\ \bottomrule
\end{tabular}
\caption{Correlations ($\rho$) between human direct assessment scores (\textbf{d}) from IWSLT23-ACL and metric/QE scores (\textbf{m} or \textbf{q}) for English-to-German speech translation.
E2E \speechQE{} scores correlate better with  human labels than cascaded approaches.
}
\label{table:cas_vs_e2e_humanda}
\end{table}

\begin{table*}[t]
\centering
\fontsize{8.5}{13}\selectfont
\begin{tabular}{l|cccc|c}
\toprule
$\rho = corr(\text{\textbf{q}}, \text{\textbf{m} or \textbf{d}})$     & \multicolumn{4}{c|}{CoVoST2 Es2En Test} & IWSLT23 \\

    & $\text{m}_{\text{xCOMET-XL}}$  & $\text{m}_{\text{xCOMET-XXL}}$      & $\text{m}_{\text{MetricX-XL}}$   & $\text{m}_{\text{MetricX-XXL}}$ & En2De $d$ \\
\midrule[\heavyrulewidth]
\arrayrulecolor{white}\hline
\multicolumn{6}{c}{\textit{\textbf{Cascaded Model with XXL Size vs E2E speech-LLM}}}      \\
$\text{q}_{cas} =$ ASR (1.5B) → xCOMET-XL-qe (3.5B)  &  0.892    & 0.800           & 0.782        & 0.788 &0.485 \\
$\text{q}_{cas} =$ ASR (1.5B) → xCOMET-XXL-qe (10.7B)& 0.787    & \textbf{0.873}   & 0.708        & 0.734 &0.486  \\
\hline\arrayrulecolor{black}
$\text{q}_{cas} =$ ASR (1.5B) → MetricX-XL-qe (3.7B)& 0.803    & 0.758   & 0.803   & 0.766   &0.495\\
$\text{q}_{cas} =$ ASR (1.5B) → MetricX-XXL-qe (13B)& 0.700    & 0.677   & 0.652   & 0.694   &0.502\\
\multicolumn{6}{c}{\textit{\textbf{Cascaded text-LLM vs E2E speech-LLM}}}      \\
$\text{q}_{cas} =$ ASR (1.5B) → \textit{text-TowerInstruct-LoRA} (7B) & 0.852    & 0.816   & 0.780        & 0.785  &\_ \\
$\text{q}_{e2e} =$ \textit{TowerInstruct-LoRA+Adapter-pt-Fixed} (7.5B)  & \textbf{0.895}    & 0.827   & \textbf{0.834}        & \textbf{0.834}   &\textbf{0.509}  \\
\bottomrule
\end{tabular}
\caption{
Impact of model size and architecture choices.
The table reports correlations ($\rho$) between \speechQE{} system scores (\textbf{q}) and either metric scores (\textbf{m}) or human direct assessment scores (\textbf{d}, right-most column).
Regardless of the size of the text-QE model, the E2E \speechQE{} system mostly outperforms the cascaded system.
Also, the cascaded system with a similar architecture of text-LLM shows lower performance than E2E \speechQE{} system.
}
\label{table:cas_vs_e2e_size}
\end{table*}




In Table~\ref{table:cas_vs_e2e_humanda}, we compare the output quality scores from \speechQE{} systems with human direct assessment (DA) scores from the IWSLT-ACL test set, instead of metric scores as in the previous sections.
We use the ASR output provided by \citet{salesky-etal-2023-evaluating}.\footnote{We tried Whisper ASR systems, but the output quality was not acceptable, likely due to the IWSLT23-ACL set being out-of-domain and covering highly technical NLP topics. The ASR provided is Azure API speech-to-text service, which we believe performs comparably to SOTA ASR models.}
Overall correlations in the IWSLT-ACL setting are lower compared to the prior section.
We hypothesize that this may be due in part to the out-of-domain nature of this test set (NLP technical talks), and to the fact that the direct assessment task performed by human judges differs from the tasks performed to obtain the gold ratings that informed our QE and metric model (MQM and WMT DA).

\paragraph{Metric vs Gold-QE.} 
The best correlation between human DA and cascaded \textQE{} with gold transcription (0.580) shows a higher coefficient than the best metric-human correlation (0.557), unlike the assumptions that metric scores would better correlate with human scores as in \citet{freitag-etal-2023-results}.
This could result from the annotation process, such as source-based DA, where annotators are shown the source text and the translated target text but not the reference text, or they are shown re-segmented translation system output along with the previous and next system outputs as described in \citet{sperber-etal-2024-evaluating-iwslt2023}.

\paragraph{E2E vs Cascaded.} 
The best E2E \speechQE{} system outperforms all ASR cascaded systems in correlation with human DA.
The ASR + WMT23-CometKiWi combination shows the highest correlation among the ASR-based configurations (0.503), but it is still slightly lower than the best E2E system (0.509). 
Notably, this best E2E system is also the top performer in the previous section. 
Overall, the data suggests that the best-practice E2E system is more effective in aligning with human judgments on translation quality compared to all cascaded systems with ASR.

\subsection{Cascaded Model Size and Architecture}
\label{sec:results_size}

Is the dominance of E2E over cascaded models due to the E2E parameter size rather than its end-to-end nature?
We address this question by varying the model size and architectural similarity between the cascaded and E2E \speechQE{} system.

\paragraph{Cascaded with XXL Size.}
In Table~\ref{table:cas_vs_e2e_size}, we evaluate cascaded systems based on bigger text-QE models---\textit{text-TowerInstruct-qe}(7B), xCOMET-XXL-qe (10.7B), and Metric-23-XXL-qe (13B))---resulting in cascaded \speechQE{} systems whose total size is bigger than that of E2E (e.g. total 14.5B of cascaded MetricXXL vs 7.5B of E2E). 
We also extend the size of metric models in the CoVoST2 comparison.
The larger text-QE system generally correlates better with human quality score than smaller cascaded system (rightmost column); however, the performance is still below that of the E2E.
Similarly in CoVoST2 test results, the E2E system outperforms the cascaded system regardless of the size of the text-QE model, except for the case where xCOMET-XXL metric favors the QE scores of the same model.

Overall, E2E models tend to show a higher correlation than the cascaded systems with similar/bigger-sized text-QE models, showing the advantages of the E2E system extend across efficiency considerations.

\paragraph{Cascaded with text-LLM.}
We LoRA fine-tune the TowerInstruct model in Spanish-to-English direction with similar training methods to E2E \speechQE{} model but only with text modality input.
This produces a text-based QE model based on the same TowerInstruct-7B model as the E2E \speechQE{} model. Pairing it with ASR results in a cascaded \speechQE{} system with 8.5B parameters as opposed to 7.5B for the E2E system. Yet, the E2E system still outperforms this version of cascaded model.
Besides the efficiency advantage, we can also conclude that the improvements are coming from the E2E nature of the approach rather than the LLM-based solution, reaffirming that E2E system is better suited for \speechQE{} task than the cascaded system.


\begin{table}[]
\centering
\fontsize{8.2}{12}\selectfont
\begin{tabular}{lccc}

\toprule
\textbf{ESD for ST}                            & Precision & Recall & F1 Score \\
\midrule[\heavyrulewidth]
\multicolumn{4}{c}{\textbf{\textit{Cascaded Systems}}}                                              \\
$\text{txt-ESD}(\text{gold }t,h)$        & 0.438     & 0.591  & 0.503    \\
$\text{txt-ESD}(\text{w-large-v2}(a),h)$ & 0.434     & 0.550  & 0.485    \\
$\text{txt-ESD}(\text{w-medium}(a),h)$   & 0.429     & $\overline{\mbox{0.540}}$  & 0.478    \\
$\text{txt-ESD}(\text{w-small}(a),h)$   & \underline{0.413}     & 0.535  & $\overline{\mbox{0.466}}$    \\
$\text{txt-ESD}(\text{w-base}(a),h)$     & 0.385     & 0.550  & 0.453    \\
\midrule[\heavyrulewidth]
\multicolumn{4}{c}{\textit{\textbf{End-to-End Systems}}}                                            \\
\textit{TowerInst-Fixed+Adt}$(a,h)$                   & 0.411     & 0.542  & 0.467   \\\bottomrule
\end{tabular} 
\caption{Zero-shot error span detection for speech translation (\speechESD{}) on CoVoST2 Spanish-to-English test. 
Even without being explicitly trained by the \speechESD{} task, E2E model performs decently suggesting that \textLLM{} ability is transferable to speech LLM in a zero-shot manner.
}
\label{table:cas_vs_e2e_esd4st}
\end{table}
\begin{table*}[t]
\centering
\fontsize{8.5}{12}\selectfont
\begin{tabular}{lll}
\toprule
\multicolumn{3}{l}{Spanish-to-English ST Example} \\\midrule
Gold transcription   & \multicolumn{2}{l}{\underline{Carpanedo} participó en dos carreras individuales del \underline{campeonato} aparte de la competencia}     \\
                       & \multicolumn{2}{l}{del miércoles.}     \\
ASR                  & \multicolumn{2}{l}{\textbf{Calpaniado} participó en dos carreras individuales del \textbf{campamento}, aparte de las competencias} \\
                     & \multicolumn{2}{l}{del miércoles.} \\
Hypothesis           & \multicolumn{2}{l}{\textbf{Calpaniado} participated in two individual races of the \textbf{camp}, apart from the Wednesday races.}                \\
Reference            & \multicolumn{2}{l}{Beyond Wednesday's event, \underline{Carpanedo} competed in two individual races at the \underline{Championships}.}                  \\
\midrule[\heavyrulewidth]
Systems              & \speechQE{} Scores & Error Span Detection                                                                                                 \\\midrule
Quality/Error Span Labels & 0.611  & \textbf{Calpaniado} -- major, of the \textbf{camp} -- major, \textit{races}--major                                                                \\
Cascaded Predictions & 0.932 & \textbf{camp}--minor, \textit{race}--minor                                                                                                \\
E2E Predictions & 0.497  & \textbf{Calpaniado} -- major, \textbf{camp} -- major                                                                                    \\\bottomrule
\end{tabular}
\caption{
Example of Spanish-to-English speech translation and quality estimations of \speechQE{} systems.
Bolded text represents the wrong ASR or ST spans while underlined indicates the correct ones.
Cascaded \speechQE{} incorrectly estimates the translation quality of the hypothesis due to speech recognition error, while E2E could correctly catch the errors in the ST.
}
\label{table:example_es2en}
\end{table*}

\subsection{Zero-Shot Error Span Detection for ST}
\label{sec:results_esd4st}

Simply providing the quality score may offer a straightforward indication of translation quality, but it can be difficult to interpret when trying to identify specific issues \citep{lu2024error}.
To broaden the scope of QE beyond overall numerical ratings, we further explore an error span detection (ESD) for ST task (\speechESD{}) that predicts the error span within the hypothesis \citep{blain-etal-2023-findings}.

We test our E2E model in a zero-shot manner where \speechESD{} is an unseen task during the speech adaptation.
Since the TowerInstruct is fine-tuned from its base model with several translation-related tasks including error span detection, 
we can see how effectively the method of injecting speech modality generalizes the capability of \textLLM{} to speech LLM without explicitly training the target speech task.
We evaluate quantitatively in this section and also qualitatively in Section~\ref{sec:results_example}.

\paragraph{Experimental Settings.} 
We use the error span output of the xCOMET metric function as reference-based error span labels and compare the E2E and cascaded system where TowerInstruct is a \textESD{} model.\footnote{We did not compare with text-xCOMET-qe in this case as we are not training \speechESD{} explicitly like \speechQE{} and xCOMET-qe output are similar to that of xCOMET-metric.} We use the same test set as \speechQE{}. 
The input of the ESD task is source and hypothesis as in the QE task.
We 
calculate the F1 score following \citet{blain-etal-2023-findings}.
For the E2E model, we only run the model that fixes the \textLLM{}, as the model performs exclusively on a few trained tasks when the weights of \textLLM{} are updated with those tasks.
Also, we build an additional \speechQE{} train set from FLEURS train set \citep{fleurs2022arxiv} and include it into a single phase \speechQE{} training to have better meaningful results in ESD, especially in qualitative analysis.

\paragraph{E2E vs Cascaded.} 
We show F1 score, recall, and precision in Table~\ref{table:cas_vs_e2e_esd4st}.
Cascaded systems show the best performance in \speechESD{} indicating that they remain the preferred choice for achieving the highest performance when we do not have speech training data for the target task.
Still, even without being explicitly trained by the \speechESD{} task, the E2E model performs decently by outperforming cascaded with medium-quality ASR in recall and cascaded with whisper-small in F1-sore.
This suggests that \textLLM{} ability is transferable to speech LLM in a zero-shot manner. 

\subsection{Example Analysis}
\label{sec:results_example}

We analyze the examples of how E2E and cascaded \speechQE{} systems score the speech translation quality and detect the error spans.
Table~\ref{table:example_es2en} shows examples of Spanish-to-English speech translation from whisper-large-v2 and quality estimations of \speechQE{} systems, where the ASR model of the cascaded system is whisper-medium.
We use xCOMET metric outputs of scores, error spans, and severity as the quality and error labels, similar to the setting of  Section~\ref{sec:results_main} and \ref{sec:results_esd4st}.

The example translation has two major errors in ``Calpaniado'' and ``camp'', which are supposed to be translated into ``Carpanedo'' and ``championship''.
However, the cascaded system estimates the quality of this translation as high as 0.93, and could not detect the error spans or its severity correctly.
These issues primarily arise because ASR incorrectly transcribed the name ``Calpaniado'' as``Calpaniado'' and the word ``campeonato'' (meaning ``championship'') as ``campamento'' (meaning ``camp'')
In contrast, E2E \speechQE{} system is not affected by these issues and correctly detects those major errors.
We discuss another example of En2De in Appendix~\ref{sec:example_en2de}.

This example shows that the E2E system is more robust to speech representation error in estimating quality and indicating the error spans for ST.

\section{Related Work}
\label{sec:related_works}

Recent work has explored how to inject additional modalities into a model pre-trained on a single modality. 
Various configurations have been proposed to meet different demands including 
speech modality into \textLLM{} \citep{wu2023decoderonly, wang2023blsp, wang2023slm}, 
visual modality into \textLLM{} \citep{liu2023visual, pmlr-v202-li23q},
visual modality into speech foundation model \citep{seo2023avformer, may2023audiovisual,han-etal-2024-xlavsr}, and 
audio-visual modalities into \textLLM{} \citep{zhang-etal-2023-video}.

When injecting the speech modality into \textLLM{}, the main challenges are aligning long speech signals to corresponding text sequences with the same semantic contents, while avoiding overfitting to default training tasks like ASR and ST.
Several methods of compressing and aligning the speech and text sequence include the use of
convolutional layer \citep{wang2023blsp},
CTC compression \citep{wu2023decoderonly, pan2023cosmic}, and
random downsampling \citep{wang2023slm}.
Many mention the problem of task overfitting to homogeneous fixed instruction training on limited tasks.
They suggest training on many diverse tasks \citep{chu2023qwenaudio, tang2024salmonn} or
tuning on diverse speech instructions with TTS-augmented instruction datasets \citep{wang2023slm,pan2023cosmic}.

However, most of these works focus on ASR, ST, QA, and general instruction following within speech comprehension tasks \citep{gaido-etal-2024-speech}. This paper initiates their application to the understudied \speechQE{} problem.

\section{Conclusion}
\label{sec:conclusion}

This work focused on the task of \speechQE{}, evaluating the quality of speech translation using both cascaded systems and end-to-end systems. 
We developed an E2E \speechQE{} model, proposing methods for corpus creation, training strategies, and architectural design. 
Our findings indicate that E2E systems are generally better suited to estimate the quality of direct speech translation. 
Additionally, we examined the error span detection task for ST finding that E2E speech model transfer ability from text-based LLM while cascaded systems with state-of-the-art ASR still hold advantages in performance.
We conclude that \speechQE{} needs dedicated attention separate from \textQE{}, due to the growing use cases of ST and the significant potential for further improvements in this field.

Quality estimation in the speech domain opens up a wide range of potential applications. In addition to the promise of helping people use speech translation systems more reliably in their daily lives, quality estimation can enhance speech translation itself, for instance by enabling prefix-to-prefix quality estimation for re-translation and simultaneous speech translation.
We contribute data, code, and models to support future work that broadens the scope of the translation-related tasks for the speech domain.

\section*{Limitations}
\label{sec:limitations}

This work assumes that we can use quality evaluation schemes designed for text translation and port them directly to speech to distill the quality estimation ability while adapting it to the speech domain.
However, some errors might matter more when translating text than when translating speech (e.g., punctuation, capitalization), while speech inputs might raise new issues (e.g., segmentation).  
In future work, we encourage the collection of quality annotations specifically designed for speech translation and look forward to investigating how to transfer knowledge from \textQE{} systems in those settings.


Our E2E models are trained with an A6000 GPU with 8 instances per batch updating up to 200k steps. 
Training with larger number of GPUs and batch size, as is often the case with speech LLM training, could show better performance in \speechQE{}.

Our training tasks include ASR, ST, and \speechQE{} with fixed instructions which interfere with the success of downstream zero-shot tasks like error span detection.
Further augmenting the training tasks with speech instruction tuning and diverse speech question answering tasks could enhance the performance of ESD.


We experimented with two language pairs, English-to-German and Spanish-to-English, both of which are European languages. We could expand language diversity in future work by including non-European languages, which would help assess the generalizability and robustness of our models across different linguistic and cultural contexts.

We have explored a single type of architecture for speech LLM.
Investigating various architectural approaches could help better understand their impact on performance and robustness in \speechQE{} performance and transferability of knowledge.

\section*{Acknowledgments}

This work was supported, in part, by the Human Language Technology Center of Excellence at Johns Hopkins University.
We also extend our gratitude to the team of the SCALE 2023 workshop on Translation of Conversational Speech, whose findings and resources gave us a headstart on this project.
Finally, we thank the anonymous reviewers, Nishant Balepur, Xinchen Yang, Dayeon Ki, and the members of the \abr{clip} lab at \abr{umd} for their insightful and constructive feedback.

\bibliography{bib/anthology_select, bib/custom, bib/speech}
\clearpage
\appendix

\begin{table}[t]
\centering
\fontsize{8.5}{13}\selectfont
\begin{tabular}{lcl}
\toprule
\textbf{Out-of-Domain} Test set (FLEURS)                & \multicolumn{2}{c}{Es2En} \\
\speechQE{} score q $\downarrow$                         & $\text{m}_{\text{xCOMET}}$      & $\text{m}_{\text{MetricX}}$     \\
\midrule[\heavyrulewidth]
\multicolumn{3}{c}{\textit{\textbf{Cascaded \speechQE{} Systems }}$\,\rho_{cas} = corr(\text{\textbf{q}}_{cas}, \text{\textbf{m}})$}              \\
$\text{xCOMET-qe}(\text{gold }t,h)$                    & 0.945       & $\overline{\mbox{0.849}}$      \\
$\text{xCOMET-qe}(\text{whspr-large-v3}(a),h)$ & 0.919       & 0.824      \\
$\text{xCOMET-qe}(\text{whspr-large-v2}(a),h)$ & 0.919       & 0.825      \\
$\text{xCOMET-qe}(\text{whspr-medium}(a),h)$      & $\overline{\mbox{0.906}}$       & 0.813      \\
$\text{xCOMET-qe}(\text{whspr-small}(a),h)$        & 0.895       & 0.804      \\
$\text{xCOMET-qe}(\text{whisper-base}(a),h)$        & 0.852       & 0.776      \\\hline
$\text{MetricX-qe}(\text{gold }t,h)$ & $\overline{\mbox{0.855}}$      & \underline{0.893}       \\
$\text{MetricX-qe}(\text{whspr-large-v3}(a),h)$                & 0.834      & 0.858       \\
$\text{MetricX-qe}(\text{whspr-large-v2}(a),h)$                & 0.833      & 0.860       \\
$\text{MetricX-qe}(\text{whspr-medium}(a),h)$                 & 0.815      & 0.840       \\
$\text{MetricX-qe}(\text{whspr-small}(a),h)$                   & 0.791      & 0.810       \\
$\text{MetricX-qe}(\text{whspr-base}(a),h)$                    & 0.709      & 0.726       \\
\midrule[\heavyrulewidth]
\multicolumn{3}{c}{\textit{\textbf{End-to-End \speechQE{} Systems }}$\,\rho_{e2e} = corr(\text{\textbf{q}}_{e2e}, \text{\textbf{m}})$}            \\
\textit{TowerInst-LoRA+Adapter-pt}$(a,h)$       & 0.897       & 0.858      \\
\textit{TowerInst-LoRA+Adt-pt-Fixed}$(a,h)$ & 0.892       & 0.849\\
\multicolumn{2}{c}{\textit{\textbf{Adding FLEURS to E2E Training \quad\quad}}}  \\   
\textit{TowerInst-LoRA+Adapter-pt}$(a,h)$       & 0.904       & 0.872      \\
\textit{TowerInst-LoRA+Adt-pt-Fixed}$(a,h)$ & \textbf{0.906}       & \textbf{0.873}\\

\bottomrule
\end{tabular}
\caption{
Correlations on out-of-domain (OOD) test set of Spanish-to-English FLEURS. 
Cascaded shows better audio domain robustness than E2E as E2E models are trained on limited data. 
Still, E2E outperforms gold-cascaded when compared with cross QE-metric cascade configuration in different model families.
We also experiment with additional FLEURS training, which increases (now in-domain) FLEURS test correlation score.
}
\label{table:cas_vs_e2e_ood}
\end{table}

\section{Robustness to Out-of-Domain Test Sets}
\label{sec:results_ood}
We also explore how the \speechQE{} systems are robust to the domain changes.
We build a test set with FLEURS \citep{fleurs2022arxiv} for out-of-domain (OOD) evaluation following the same protocol as an in-domain test set.
Table~\ref{table:cas_vs_e2e_ood} shows correlations between \speechQE{} system score and metric score on the out-of-domain test set of FLEURS.

\paragraph{Effect of ASR quality in Cascaded.} 
We present cascaded results with a wide quality range of ASR, from whisper-large v3 to whisper-base. 
The correlations are proportional to the ASR performances, while gold cascaded is an upper bound.

\paragraph{Robustness Effect of Training E2E Adapter with Target Task}
In contrast to Section~\ref{sec:results_main}, the best-performing E2E model is the model that updates the pre-trained adapter weight in the final training stage with the \speechQE{} task.
We note that the training of the adapter and the final E2E model is based solely on Common Voice audio, where the adapter is trained with ASR and ST tasks and the final E2E model is only trained with the \speechQE{}.

We conclude that E2E models become more robust to audio domain shift if the speech adapter is trained with the target task---\speechQE{} in this case---instead of being frozen.

\paragraph{E2E vs Cascaded.} 
The results suggest that cascaded systems have better domain robustness when comparing the correlation between matching QE and metric models like the pair of ASR + xCOMET-qe and xCOMET metric scores.
In those cases, the E2E system (e.g. 0.858 in MetricX) only outperforms the cascaded system with medium-quality ASR systems (e.g. 0.840 with whisper-medium ASR).
This advantage is likely due to  ASR systems being trained on a broader domain of audio corpora, whereas E2E systems are limited to Common Voice domain.
Nevertheless, the E2E system shows competitive correlations in settings with non-matching QE and metric models (e.g., xCOMET-qe and MetricX metric), outperforming the cascaded systems of gold transcription and \textQE{}.


\begin{table*}[t]
\centering
\fontsize{9}{13}\selectfont
\begin{tabular}{lcccc|c}
\toprule
$\rho = corr(\text{\textbf{q}}, \text{\textbf{m} or \textbf{d}})$     & \multicolumn{2}{c}{CoVoST2 Es2En} & \multicolumn{2}{c|}{CoVoST2 En2De}  & IWSLT23 \\

    & $\text{m}_{\text{xCOMET}}$      & $\text{m}_{\text{MetricX}}$ & $\text{m}_{\text{xCOMET}}$    & $\text{m}_{\text{MetricX}}$ & En2De $d$ \\
\midrule[\heavyrulewidth]
\multicolumn{6}{r}{\textit{\textbf{Cascaded \speechQE{} Systems Correlations}}$\,\rho_{cas} = corr(\text{\textbf{q}}_{cas}, \text{\textbf{m}}) \quad\quad\quad\quad\quad \rho_{cas} = corr(\text{\textbf{q}}_{cas}, \text{\textbf{d}})$   }   \\     
$\text{q}_{cas} = \text{xCOMET-qe}(\text{ASR}(a),h)$  & 0.892        & 0.782   & 0.910        & 0.821 &0.485 \\
$\text{q}_{cas} = \text{MetricX-qe}(\text{ASR}(a),h)$ & 0.803        & 0.803   & 0.854        & 0.871 &0.495  \\
\midrule[\heavyrulewidth]
\multicolumn{6}{r}{\textit{\textbf{End-to-End \speechQE{} Systems Correlations}}$\,\rho_{e2e} = corr(\text{\textbf{q}}_{e2e}, \text{\textbf{m}}) \quad\quad\quad\quad\quad \rho_{e2e} = corr(\text{\textbf{q}}_{e2e}, \text{\textbf{d}})$   }   \\    
$\text{q}_{e2e} = \text{\textit{TowerInstruct-LoRA+Adapter-pt}}(a,h)$  & 0.890  & 0.833  & 0.922  & 0.872  & 0.492 \\
$\text{q}_{e2e} = \text{\textit{TowerInstruct-LoRA+Adapter-pt-Fixed}}(a,h)$  & \textbf{0.895}        & \textbf{0.834}   & \textbf{0.925}        & \textbf{0.873}   & 0.5085 \\
\multicolumn{2}{c}{\textit{\textbf{Adding FLEURS to E2E Training}}}  \\  
$\text{q}_{e2e} = \text{\textit{TowerInstruct-LoRA+Adapter-pt}}(a,h)$ & 0.893        & 0.828   & 0.916        & 0.868  &0.501 \\
$\text{q}_{e2e} = \text{\textit{TowerInstruct-LoRA+Adapter-pt-Fixed}}(a,h)$  & 0.888        & 0.826   & 0.920        & 0.871 &\textbf{0.5091} \\
\bottomrule
\end{tabular}
\caption{
CoVoST2 and IWSLT23-ACL results of the E2E models trained on a single-domain of CoVoST2 corpus (first two rows of E2E section) and multi-domain corpus including CoVoST2 and FLEURS (last two rows).
Adding the FLEURS domain decreases performance on the CoVoST2 domain but slightly improves in correlation with IWSLT23-ACL human direct assessment scores, while still outperforming the cascaded \speechQE{} system.
}
\label{table:cas_vs_e2e_fleurs_train}
\end{table*}
\section{Adding FLEURS set to E2E Training}
\label{sec:results_add_fleurs_train}
Training a model on a single speech domain may lead to learning domain-specific speech representation, such as particular accents or speaking styles. 
We experiment with an additional \speechQE{} training set to verify whether the conclusion from single-domain experiments holds in broader settings.
We create an additional \speechQE{} training set from the FLEURS dataset (20k), which is relatively small compared to CoVoST2 (more than 500k).
We include it into a single phase \speechQE{} training, which is the same corpus setting described in Section~\ref{sec:results_esd4st}.
We present the evaluation results on CoVoST2 and IWSLT23-ACL in Table~\ref{table:cas_vs_e2e_fleurs_train} and on FLEURS in Table~\ref{table:cas_vs_e2e_ood}, specifically in the last two rows of each table.

First, adding the FLEURS domain shows higher correlations on the FLEURS domain as anticipated (last two rows of Table~\ref{table:cas_vs_e2e_ood}).
In contrast, it reduces performance on the CoVoST2 domain but still outperforms the cascaded \speechQE{} systems (Table~\ref{table:cas_vs_e2e_fleurs_train}).
Interestingly, the correlation between the human score of IWSLT-ACL and the \speechQE{} system score (rightmost column in Table~\ref{table:cas_vs_e2e_fleurs_train}) shows that adding even a small set from another domain slightly increases the alignment with human judgments.
Although this improvement may not be statistically significant, it suggests that training on multiple speech domains (CoVoST2 + FLEURS) increases robustness against domain shifts during testing (as IWSLT ACL is also out-of-domain).

In conclusion, the findings from single-domain experiments remain valid after incorporating the FLEURS set into training, while also indicating increased robustness to domain shifts.

\section{Additional Examples in En2De}
\label{sec:example_en2de}
Table~\ref{table:example_en2de} shows examples of English-to-German speech translation results from s2t-medium-mustc-multilingual-st in Table~\ref{table:bleu}.
The translation has several major errors and both cascaded and E2E systems are able to detect the errors.
However, the cascaded system incorrectly predicts the severities as minor and ends up estimating the quality score to be 0.852.
One could be partly due to an ASR error where it incorrectly transcribed ``GBP'' as ``GPP'', which might trigger the cascaded system to set its severity as a minor for the translation of ``GP''.

\section{Additional Experimental Details}
\label{sec:add_exp_detail}
For E2E training, we use a learning rate of 5e-5 and a weight decay of 0.05.
For LoRA training, we update q|k|v|o projection in each attention layer with the rank of $r=16$ and a scaling parameter of $\alpha=32$.
The size of the resulting E2E \speechQE{} model is about 8.5B given that TowerInstruct \textLLM{} is 7B and whisper-large-v2 is 1.5B.
For decoding, we use a temperature of 0.1 and set the maximum new tokens up to 500.
The presented numbers in all tables are a single run for cascaded where the outputs do not change with the same input and the mean of three runs for E2E.
We use off-the-shelf models from the huggingface hub and use \texttt{torch} and \texttt{transformer} libraries for the implementation.

\section{Discussion of the Task Terminology}
\label{sec:task_name}
In the research area of machine translation (MT), the term QE traditionally stands for machine translation quality estimation, though the more precise acronym is MTQE. Also, MT typically indicates text-to-text translation, while ST refers to speech-to-text translation. Given the implications of QE, we add ``speech'' to indicate the task of quality estimation for \textbf{speech} translation, where the more accurate acronym would be STQE. 
We use \speechQE{} for speech translation quality estimation and \textQE{} for machine translation quality estimation as main wordings instead of (more accurate) alternatives of STQE and MTQE to emphasize the contrast between speech and text and to facilitate easier reading. While \speechQE{} could be ambiguous considering that it can be QE either for ASR or ST, previous works on ASR quality estimation \citep{negri-etal-2014-quality, rubenstein2023audiopalm} use the phrase ``ASR-QE'', which safely distinguishes them from STQE or \speechQE{}.


\begin{table*}[t]
\centering
\fontsize{8.5}{12}\selectfont
\begin{tabular}{lll}
\toprule
\multicolumn{3}{l}{English-to-German ST Example} \\\midrule
Gold transcription   & \multicolumn{2}{l}{The official Falklands currency is the Falkland pound (FKP)}     \\
                       & \multicolumn{2}{l}{whose value is set equivalent to that of one British pound (GBP).}     \\
ASR                  & \multicolumn{2}{l}{The official Falklands currency is the Falkland Pound, FKP,} \\
                       & \multicolumn{2}{l}{whose value is equivalent to that of a British Pound, \underline{GPP}.}     \\
Hypothesis           & \multicolumn{2}{l}{Die offizielle Fäklins Währung ist ein Fäklin Pfund, FKP,}                \\
                       & \multicolumn{2}{l}{der uns wertvoll ist, genauso wie ein britischer Pfund, \textbf{GP}.}     \\
Reference            & \multicolumn{2}{l}{Die offizielle Währung der Falklandinseln ist das Falkland Pound (FKP),}                  \\
                       & \multicolumn{2}{l}{dessen Wert in Einklang mit dem Wert des Britischen Pfunds (GBP) festgelegt wird.}     \\
\midrule[\heavyrulewidth]
Systems              & \speechQE{} Scores & Error Span Detection                                                                                                 \\\midrule
Quality/Error Span Labels & 0.539  & ``e Fäklins W'' -- major, ``hrung ist ein Fäklin Pfund, FKP,                                                                \\
                       & &der uns wertvoll ist, genauso wie ein britischer Pfund, \textbf{GP}.'' -- major,     \\
Cascaded Predictions & 0.852   & ``e Fäklins Währung'' -- minor, ``ein Fäklin Pfund'' -- minor, FKP -- minor, \\
                       & &  ``der uns wertvoll ist, genauso'' -- minor, ``britischer Pfund, \textbf{GP}'' -- minor     \\
E2E Predictions & 0.550  & Fäklins -- major, FKP -- major, ``uns wertvoll ist'' -- major,                                                                                     \\
                       & &  ``genauso wie'' -- major, ``britischer Pfund -- major, \textbf{GP} -- major     \\\bottomrule
\end{tabular}
\caption{
Example of English-to-German speech translation and quality estimations of \speechQE{} systems.
Both cascaded and E2E \speechQE{} systems could detect errors. 
However, the cascaded system estimates the severity lower than that of the metric labels partly due to ASR error while E2E could estimate the quality closely to labels. 
}
\label{table:example_en2de}
\end{table*}
\begin{figure*}[t]
    \centering
    \includegraphics[width=0.95\textwidth]{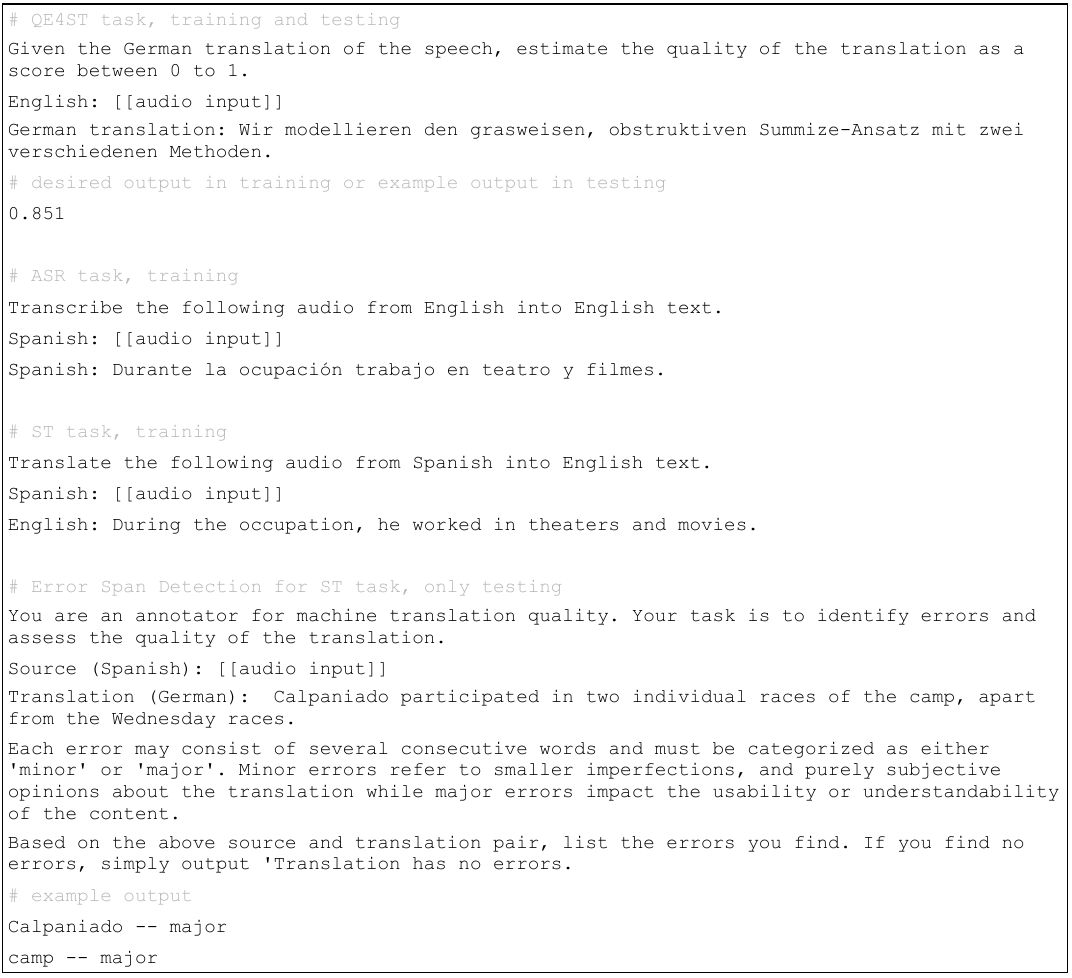} 
    \caption{Prompt template of \speechQE{} (quality estimation for speech translation), ASR, ST, and \speechESD{} (error span detection for ST) task.
    }
    \label{fig:prompt_exmaple}
\end{figure*}

\end{document}